\pdfoutput=1

\documentclass[11pt]{article}

\usepackage[preprint]{acl}

\usepackage{times}
\usepackage{latexsym}

\usepackage[T1]{fontenc}

\usepackage[utf8]{inputenc}

\usepackage{microtype}

\usepackage{inconsolata}

\usepackage{graphicx}

\usepackage{colortbl}
\usepackage{xcolor}

\usepackage{booktabs}
\usepackage{algorithm}
\usepackage{algorithmic}
\usepackage{amsmath}
\usepackage{comment}
\usepackage[capitalize,noabbrev]{cleveref}

\title{Zero-Shot Verification-guided Chain of Thoughts}

\author{Jishnu Ray Chowdhury\thanks{$\quad$ The main work was done while the first author was a PhD Student at University of Illinois at Chicago. The work is unaffiliated with Bloomberg.} \\
  University of Illinois at Chicago \\
  \texttt{jishnu.ray.c@gmail.com} \\\And
  Cornelia Caragea \\
  University of Illinois at Chicago \\
  \texttt{cornelia@uic.edu} \\}

\begin{document}
\maketitle
\begin{abstract}
Previous works have demonstrated the effectiveness of Chain-of-Thought (COT) prompts and verifiers in guiding Large Language Models (LLMs) through the space of reasoning. However, most such studies either use a fine-tuned verifier or rely on manually handcrafted few-shot examples. In contrast, in this paper, we focus on LLM-based self-verification of self-generated reasoning steps via COT prompts in a completely {\em zero-shot regime}. To explore this setting, we design a new zero-shot prompt, which we call COT STEP, to aid zero-shot decomposition of reasoning steps and design two new zero-shot prompts for LLM-based verifiers. We evaluate the verifiers' ability to classify the correctness of reasoning chains and explore different ways to use verifier scores in guiding reasoning for various mathematical and commonsense reasoning tasks with different LLMs. 
\end{abstract}
\section{Introduction}
Large Language Models (LLMs) \cite{dubey2024llama,brown2020language} have revolutionized the field of NLP by enabling state-of-the-art (SOTA) performance in several tasks merely by smart prompting \cite{wei2022emergent,bubeck2023sparks}. A critical landmark in the art of prompting for multi-step reasoning tasks is the Chain-of-Thought (COT) prompting  \cite{nye2022show, wei2022chain}, which elicits LLMs to generate step-by-step reasoning chains before providing the answer to a given question. 
Building upon COT prompting, several recent works incorporate a verifier mechanism to improve LLMs' performance. 
For instance, \citet{cobbe2021training,li2023making,weng-etal-2023-large} use a verifier to evaluate the correctness and score each reasoning step, whereas \citet{gandhi2023strategic,yao2023tree,hao2023reasoning,hao2024llm} use a verifier's scores to do a tree-search over the space of reasoning steps. 

However, these works have several drawbacks including: 
1. most of them rely on few-shot prompting, i.e., they rely on exemplars (often manually handcrafted) of how the task is to be performed. This selection of the exemplars brings additional challenges as to which exemplars should be selected, how many, etc; 2. The verifier is also typically either fine-tuned on labeled datasets or rely on further verification-related few-shot exemplars. Properly fine-tuning a verifier may require a large number of human annotations which may be impractical in different tasks and domains.
In contrast, in this paper, we focus on LLM-based self-verification of self-generated reasoning steps via COT prompts in a completely {\em zero-shot regime}. To achieve this, we 
propose a new zero-shot COT-based prompt, called COT STEP, for generating reasoning steps (i.e., to aid zero-shot decomposition of reasoning steps) and two new zero-shot verification prompts for using LLMs as verifiers to verify their own reasoning steps (i.e., evaluate whether the steps are correct or not). 
The zero-shot regime is valuable because in this setup, no manually selected or handcrafted examples are required. Moreover, in the complete zero-shot setting that we explore, there is no reliance on any additional data resource for exemplar construction. Furthermore, a general purpose zero-shot prompt can be used to set up a convenient model interface for users to request different kinds of reasoning tasks without having to worry about providing task-specific exemplars. 

In exploring the zero-shot regime, we ask three research questions (RQ1-3) in this paper.

\noindent \textbf{RQ1:} How do different zero-shot COT prompts compare against each other across different LLMs? 

\noindent \textbf{RQ2:} How well do LLM-based verifiers perform in evaluating the correctness of reasoning steps?

\noindent \textbf{RQ3:} How well can LLM-based verifier scores be used in selecting the right answer among many or in guiding reasoning in real-time? 

We address each of the questions in turn - RQ1 in $\S$\ref{sec:RQ1}, RQ2 in $\S$\ref{sec:RQ2}, and RQ3 in $\S$\ref{sec:RQ3} and evaluate them on various mathematical and commonsense reasoning tasks with different LLMs. While investigating RQ1, we find that our COT STEP prompt keep up competitively with other prompts alongside enabling automatic step decomposition. While investigating RQ2, we find that a zero-shot COT-variant of verification prompt tends to perform reasonably in verifying the correctness of mathematical reasoning steps. While investigating RQ3, we do find a noticeable benefit from using the verifier score in a step-level greedy search but the benefit vanishes when self-consistency \cite{wang2023selfconsistency} is used.   

\section{Main Components}
In this section, we discuss the main components of our experiments that are relevant for all our research questions (RQ1-3). 

\subsection{Models}
For our experiments, we primarily focus on two openly-licensed models in the main paper - (1) SOLAR (SOLAR 10.7B Instruct)\footnote{SOLAR is an instruction-tuned depth-scaled variation of Llama 2.} \cite{kim-etal-2024-solar} and (2) Phi3 (Phi 3 Mini 128K Instruct) \cite{abdin2024phi} and show additional experiments on other models such as Mistral (Mistral 7B Instruct v0.2) \cite{jiang2023mistral} and Llama3 (LLama 3.0 8B Instruct) \cite{dubey2024llama} in Appendix \ref{sec:additional}.  

\begin{table*}[t]
\small
\centering
\def\arraystretch{1.2} 
\begin{tabular}{  l | r r r r r | r r r r r} 
\toprule
\textbf{Prompt Style} & \textbf{GSM8K} & \textbf{GSMH} & \textbf{AQuA} & \textbf{StratQA} & \textbf{CSQA} & \textbf{GSM8K} & \textbf{GSMH} & \textbf{AQuA} & \textbf{StratQA} & \textbf{CSQA}\\
\hline
\multicolumn{6}{c|}{\textbf{SOLAR 10.7B Instruct}} & \multicolumn{5}{c}{\textbf{SOLAR 10.7B Instruct + SC@20}}\\
\hline
Base & 17.06 & 5.38 & 6.69 & 31.09 & 63.96 & 17.67 & 7.28 & 30.71 & 65.42 & 72.40\\
COT & \textbf{73.62} & 28.96 & 6.69 & 58.47 & 66.67 & 83.40 & 37.38 & 42.9 & 73.10 &  78.79\\
PS+ & 68.39 & 26.61 & 10.63 & 30.87 & 51.27 & 84.9 & \textbf{39.20} & \textbf{47.24} & 69.78 & 80.43\\
TAB COT & 65.2 & 26.69 & 8.66 & 52.66 & 65.19 & 80.52 & 36.16 & 41.73 & \textbf{73.89} & \textbf{80.51}\\
COT STEP & 73.54 & \textbf{30.63} & \textbf{11.02} & \textbf{58.95} & \textbf{67.32} & \textbf{84.91} & 37.45 & 38.58 & 73.76 & 79.85\\
\hline
\multicolumn{6}{c|}{\textbf{Phi3 Mini 128K Instruct}} & \multicolumn{5}{c}{\textbf{Phi3 Mini 128K Instruct + SC@20}}\\
\hline
Base & 51.40 & 24.19 & \textbf{19.29} & 9.2 & 69.53 & 77.64 & 38.29 & 45.67 & 40.13 & 73.55\\
COT & \textbf{84.84} & 37.6 & 18.89 & 44.63 & 69.37 & \textbf{91.28} & 45.03 & 64.57 & 61.75 & 76.33\\
PS+ & 81.5 & 36.24 & 14.96 & 17.21 & 56.76 & 90.9 & \textbf{46.02} & \textbf{66.54} & 55.72 & 75.18\\
TAB COT & 73.77 & 32.98 &  12.6 & 45.98 & 67.90 & 86.81 & 44.05 & 57.87 & \textbf{63.67} & 74.86\\
COT STEP & 83.25 & \textbf{38.59} & 15.75 & \textbf{51.0} & \textbf{71.75} & 90.75 & 45.49 & 65.75 & 63.49 & \textbf{77.07}\\
\bottomrule
\end{tabular}
\caption{Comparison of accuracy from different prompting strategies across different datasets. GSMH represents GSM-HARD, StratQA represents StrategyQA, and CSQA represents CommonsenseQA. SC@20 represents Self-consistency with majority voting of 20 sample generations. We bold the best result per block per dataset.}
\label{table:basic-results}
\end{table*}

\subsection{Existing Prompts}
\label{sec:existing-prompts}
In our experiments, we consider four major existing zero-shot prompt strategies. We describe them here. Note that we use a simplified representation of user-assistant conversational turns to describe the prompt structure for generality and brevity.\footnote{Different models based on how they were instruction-tuned, require different kinds of templates to represent user-assistant conversational turns which we use accordingly in experiments. The specific template information can be found in Huggingface \cite{wolf-etal-2020-transformers} tokenizer configs in the corresponding model pages.}  

\noindent \textbf{(P1) Base:} Base represents the baseline zero-shot prompting style for question answering that we use. Given a question, such as ``What is 2+2?", the prompt to elicit an answer would look like: {\color{cyan!50!black}{``User: Q: What is 2+2? \textbackslash n Assistant: A:"}}. 

\noindent \textbf{(P2) COT:} COT (Chain Of Thoughts) represents the standard zero-shot COT prompt \cite{kojima2022large}. This can be expressed as {\color{cyan!50!black}{``User: Q: What is 2+2? \textbackslash n Assistant: Let's think step by step."}}. Such a prompt is designed to prime the models to generate more explicit chain of reasoning steps before producing an answer.    

\noindent \textbf{(P3) PS+:} PS+ represents the plan-and-solve zero shot PS+ prompt presented by \citet{wang-etal-2023-plan} as an extension over plain COT. This can be expressed as: {\color{cyan!50!black}{``User: Q: What is 2+2? \textbackslash n Assistant: A: Let's first understand the problem, extract relevant variables and their corresponding numerals, and devise a plan. Then, let's carry out the plan, calculate intermediate variables (pay attention to correct numeral calculation and commonsense), solve the problem step by step, and show the answer."}}

\noindent \textbf{(P4) TAB COT:} TAB COT represents another zero shot prompt originally presented by \citet{ziqi-lu-2023tab}. This can be expressed as: {\color{cyan!50!black}{``User: Q: What is 2+2? \textbackslash n Assistant: A: \textbackslash n$|$ step $|$ subquestion $|$ process $|$ result\textbackslash n"}}. Such a prompt primes the model to perform chain of thought style reasoning in a tabular format.

Note that we mainly focus on prompt strategies that do not require any additional data (label-free or not), pseudo-data generation, or involvement in any sophisticated meta-prompting \cite{yuan2024instanceadaptive,jin2024zero,zhou2024self,chen-etal-2023-self,wan-etal-2023-universal,wan-etal-2023-better}. Such works can be explored in future work. Besides the prompts discussed above (P1-4), \citet{xu-etal-2024-reading} proposed a zero-shot prompt for re-reading the question and \citet{kong2024better} recently proposed role-play-based zero-shot prompting. They can be orthogonal augmentations to any of the existing zero-shot prompts. We do not focus on them for this paper. 

\subsection{Our COT STEP Prompt for Thought Decomposition}
\label{sec:cotstep}
Many of our experiments require LLM-based self-verification of the correctness of the reasoning steps induced by COT-styled prompts. However, this requires some strategy to decompose the reasoning steps (or thoughts) to be evaluated separately. In case of few-shot COT prompts, this can be done by providing exemplars with a specific reasoning format that can be exploited to automatically parse reasoning steps. However, that is not possible with zero-shot COT prompts. Most of the existing zero-shot COT prompts discussed above, do not provide any specific formatting constraint that would allow us to systematically demarcate reasoning steps.\footnote{We can use new lines to decompose reasoning steps but that is a non-ideal heuristic that does not guarantee a semantically meaningful chunking of steps.}  

TAB COT is an interesting exception compared to other existing prompts since it does induce the model to follow a specific tabular structure corresponding to the given columns. Thus, we can treat each generated row in the table (based on TAB COT) as a single reasoning step which can be decomposed based on new line markers (``\textbackslash n"). 
Nevertheless, we generally find the performance of TAB COT to be lower compared to other existing prompts as we will show later. We also find just the basic COT prompt to be a good enough contender ($\S$\ref{sec:basic-results}). As such, we design a simple variant of the standard zero-shot COT prompt that makes the model always generate thoughts automatically in a structured manner where each thought (reasoning step) is separated cleanly into different numbered steps. We call this prompt as COT STEP. 

\noindent \textbf{(P5) COT STEP:} COT STEP is a novel variant of zero-shot COT prompt. This can be expressed as: {\color{cyan!50!black}{``User: Q: What is 2+2? \textbackslash n Assistant: A: Let's think step by step.\textbackslash n\textbackslash nStep 1: "}}. This prompt primes the model to organize its reasoning steps into numbered steps (Step 1, Step 2, \dots). 
Given such structured outputs from COT-STEP, the reasoning steps (or ``thoughts") can easily be decomposed automatically by using `Step <number>:' as a delimiter. 

\colorlet{c1}{blue!30!white}

\begin{figure*}[ht]
\scriptsize
\fbox{\parbox{0.985\textwidth}{
\textbf{User:} 

Here is a question:

\colorbox{c1}{[QUESTION]}
\vspace{1mm}

For the given question, here is a (potentially partial) reasoning process:
\vspace{1mm}

\colorbox{c1}{[THE REASONING CHAIN SO FAR]}
\vspace{1mm}

Double-check the last reasoning step (STEP \colorbox{c1}{[NUMBER]}), let’s analyze its correctness to choose the correct option for the following question.

Question: Is the last solution (in STEP \colorbox{c1}{[NUMBER]}) correct?

(A) Yes

(B) No
\vspace{1mm}

\textbf{Assistant:} Answer: (
}}
\vspace{-1mm}
\caption{\small The {\bf R-prompt} template used for LLMs to verify/evaluate the last (potentially partial) reasoning step in a given generated reasoning chain. The blue color represents placeholders for the template that is to be filled with the input data-specific information (the question, and generated reasoning step with the last step number).}
\label{fig:eval_prompt}
\end{figure*}

\subsection{Datasets}

We evaluate our research questions on five Qusetion Answering-based reasoning datasets - (1) GSM8K \cite{cobbe2021training} (2) GSM-HARD (GSMH) \cite{gao2023pal} (3) AQuA \cite{ling-etal-2017-program} (4) StrategyQA (StratQA) \cite{geva-etal-2021-aristotle} (5) CommonsenseQA (CSQA) \cite{talmor-etal-2019-commonsenseqa}, which are mathematical reasoning (first three) and commonsense reasoning datasets (last two). 


\section{RQ1: COT Prompt Comparisons}
\label{sec:RQ1}
In this section, we seek to compare the existing zero-shot prompts alongside our COT STEP prompt. We want to ensure that COT STEP maintains reasonable performance compared to other zero-shot prompts. Moreover, some prompts like PS+ are only evaluated for a single LLM, and typically not an openly-licensed LLMs. Thus, we 
compare the zero-shot prompts under the same settings across two openly-licensed models - SOLAR and Phi3, and show results with additional LLMs in Appendix \ref{sec:additional}. Furthermore, self-consistency \cite{wang2023selfconsistency} has shown to boost the performance of any prompt and we explore it in our setting. 

\paragraph{Self-Consistency} Self-Consistency \cite{wang2023selfconsistency} 
creates multiple sample generations from a model given any prompt. Each sample is associated with an answer to the given question. After that, majority voting is applied to select the most frequent answer among the samples. 
We compare all the five prompts both with and without self-consistency. When using Self-Consistency, we use a temperature of $0.8$ and draw $20$ samples.

\subsection{Results}
\label{sec:basic-results}
We present our experiments in Table \ref{table:basic-results}. The Base (Non-COT) prompt when using Phi3 without self-consistency excelled in AQuA and is otherwise competitive in CSQA where COT-style reasoning may not be as crucial given that it is a commonsense reasoning task. Besides that, it is the worst performer with or without using self-consistency. Surprisingly, PS+ and TAB COT (although they were shown to be better than original zero-shot COT \cite{kojima2022large}) do not consistently outperform COT. Rather, without self-consistency, they often seem to perform significantly worse. Thus, these prompts may only work better compared to COT for specific models like GPT 3.5 Turbo, Davinci and such - but not generally across most models. Interestingly, when using self-consistency, these same prompts tend to become much more competitive and the results can flip. For example, PS+ on SOLAR performs the second worst on GSMH without self-consistency, but performs the best on the same task with self-consistency. Our COT STEP keeps up fairly well with other COT prompts. Without self-consistency COT STEP is even the best performer in most cases for SOLAR. With self-consistency COT STEP is still nearby to the best performer. Thus, COT STEP offers a good way to get structured outputs with cleanly separated reasoning steps in a zero-shot manner while maintaining competitive performance against most of the existing zero-shot COT prompts. We provide examples of generations from various prompts in Appendix \ref{sec:gen-examples}.

\section{RQ2: Evaluating Self-Verification}
\label{sec:RQ2}
Self-verification refers to the idea of using LLMs to verify the correctness of their own generated reasoning steps. Roughly, this is typically done by sending LLM-generated reasoning steps to the same LLM as an input alongside some appropriate prompt to evaluate those steps. The LLMs' response and its corresponding probabilities can be then used to get a ``score'' for the correctness of the generated reasoning steps that were sent as the input. In this paper, we evaluate each reasoning step separately (given the context of past reasoning steps) and assign each reasoning step with the evaluated score. Below, we describe our specific methods for obtaining the evaluation score.   

\subsection{Self-Verification: Task Setup and Prompts}
\label{sec:selfver}
{\bf Task Setup} \quad For self-verification, given an input question and its associated (potentially partial) reasoning chain, the task is to determine whether the {\bf last step} in the chain is correct or not.\footnote{We can also evaluate other dimensions such as helpfulness and factual-grounded-ness but in this paper, we focus on correctness.} Thus, the self-verification happen at a {\em step-level granularity}. 

\vspace{1mm}
\noindent {\bf R-prompt} \quad For this task, we design a zero-shot self-verification prompt as shown in Figure \ref{fig:eval_prompt}, which we call {R-prompt}. The prompt is adapted from the one-shot verification prompt used in \citet{ling2023deductive} for a simplified zero-shot context. 

\vspace{1mm}
\noindent {\bf COTR-prompt} \quad Similar to \citet{ling2023deductive}, we also explore a COT variant of the R-prompt which we call the {COTR-prompt}. After the {\color{cyan!50!black}{``Assistant: "}} in R-Prompt in Figure \ref{fig:eval_prompt}, the COTR-prompt variant replaces {\color{cyan!50!black}{``Answer: ("}} with the following line: {\color{cyan!50!black}{``Answer: Let's double-check step by step.\textbackslash n"}}. 

\subsubsection{Verification Answer Extraction}
\label{sec:eval-answer-extraction}
From Figure \ref{fig:eval_prompt}, ``(A)" represents ``Yes" to the correctness of the given reasoning step whereas ``(B)" represents ``No". Thus, as a heuristic, we then check if "A)" is present in the generated response or not. If present, we consider that the LLM predicts the last (potentially partial) reasoning step to be correct, otherwise we consider that it predicts the last step to be incorrect. Given that the R-prompt is primed to immediately generate ``A" or ``B" as an option in the next step, we restrict its maximum generation length to $2$. For COTR-prompt, we set the maximum generation length to $512$. 

\subsubsection{Verification Score Generation}
\label{sec:score-gen}
Above we discussed how we get the LLM-based self-verification prediction. Here, we discuss how we get a score for the correctness (the answer ``(A)") of the last reasoning step from the LLM. There are two components (C1 and C2) to generating an overall scalar score for a reasoning step. 

\noindent \textbf{C1: Reasoning Step Generation} The component C1 returns the average log-probability (token-wise) for the generation of the last reasoning step in the input reasoning chain (that is to be verified). Let us call the returned score (obtained during the generation with COT prompts (from \S\ref{sec:existing-prompts} and \S\ref{sec:cotstep}), e.g., with COT STEP) as $s_{C1}$. Note C1 has nothing to do with the self-verification prompt (R-prompt or COTR-prompt) or verification prediction.

\noindent \textbf{C2: Reasoning Step Verification} \quad The component C2 first calculates the average log-probability (token-wise) of the generation of the response to the verification prompt (R-prompt or COTR-prompt from \S\ref{sec:selfver}). Let us call the returned score as $\widehat{s}_{C2}$. The final value returned by C2 is then:
\begin{equation}
\vspace{-1mm}
    s_{c2} = 
    \begin{cases}
        \widehat{s}_{C2} & \text{prediction = (A)}\\
        \log(1-\exp\widehat{s}_{C2}) & \text{prediction = (B)}
    \end{cases}
\end{equation}
This scoring is reliant on the prediction (answer extraction) from the verification prompt (R-prompt/COTR-prompt) as described in \S\ref{sec:eval-answer-extraction}.
$s_{c2}$ is designed to correlate positively with the probability of correctness of the given reasoning step. 

\vspace{1mm}
\noindent \textbf{Unification:} The two scores $s_{C1}$ and $s_{C2}$ are then averaged and exponentiated to return the final unified scalar score ($s_{f}$):
\begin{equation}
    s_f = \exp \left ( \frac{s_{C1} + s_{C2}}{2} \right )
    \label{eq2}
\end{equation}

\noindent We take inspiration from \citet{xie2023self-evaluation} in incorporating both the generation probability of the reasoning step and the probability of verification prediction for the final score of the reasoning step. However, unlike \citet{xie2023self-evaluation}, our approach is simpler in that we take a simple average instead of introducing new tunable hyperparameters. In Appendix \ref{sec:additional}, we also discuss some ablation experiments where we use only C2 for scoring. There, we show that C1 is required for the best performance.

\begin{table*}[t]
\centering
\small
\def\arraystretch{1.2} 
\begin{tabular}{  l | l l l l l | l l l l l} 
\toprule
 & \multicolumn{5}{c|}{\textbf{Question Source: GSM8K}} & \multicolumn{5}{c}{\textbf{Question Source: CSQA}}\\
\textbf{Verifier Setup} & \textbf{Acc} & \textbf{TNR} & \textbf{TPR} & \textbf{Prec} & \textbf{F$_1$} & \textbf{Acc} & \textbf{TNR} & \textbf{TPR} & \textbf{Prec} & \textbf{F$_1$}\\
\hline
\multicolumn{11}{l}{\textbf{Reasoning Chain Generation Prompt: TAB COT}}\\
\hline
SOLAR + R-prompt & 57.15 & 16.9 & \textbf{97.40} & 53.96 & 69.45 & 52.85 & 9.6 & 96.1 & 51.53 & \textbf{67.09}\\
\rowcolor{lightgray!40}SOLAR + COTR-prompt & 67.30 & 49.8 & 84.8 & 62.82 & 72.17 & \textbf{53.55} & 30.60 & 76.5 & \textbf{52.43} & 62.22\\
Phi3 + R-prompt & 61.35 & 28.7 & 94.0 & 56.87 & 70.86 & 49.4 & \textbf{56.90} & 41.9 & 49.29 & 45.30\\
\rowcolor{lightgray!40}Phi3 + COTR-prompt & \textbf{80.2} & \textbf{73.5} & 86.9 & \textbf{76.63} & \textbf{81.44} & 51.95 & 15.0 & \textbf{88.9} & 51.12 & 64.91\\
\hline
\multicolumn{11}{l}{\textbf{Reasoning Chain Generation Prompt: COT STEP}}\\
\hline
SOLAR + R-prompt & 55.75 & 12.7 & \textbf{98.8} & 53.09 & 69.07 & 50.2 & 1.7 & \textbf{98.7} & 50.10 & \textbf{66.47}\\
\rowcolor{lightgray!40}SOLAR + COTR-prompt & \textbf{83.90} & 74.7 & 93.10 & \textbf{78.63} & \textbf{85.26} & 49.5 & 29.80 & 69.20 & 49.64 & 57.81\\
Phi3 + R-prompt & 59.8 & 22.7 & 96.90 & 55.63 & 70.68 & 45.4 & \textbf{57.50} & 33.30 & 43.93 & 37.88\\
\rowcolor{lightgray!40}Phi3 + COTR-prompt & 75.2 & 60.20 & 90.2 & 69.40 & 78.44 & \textbf{50.4} & 10.2 & 90.60 & \textbf{50.22} & 64.62\\
\bottomrule
\end{tabular}
\caption{Performance of SOLAR and Phi3 on our four correctness classification datasets - 1. (COT STEP, GSM8K), 2. (COT STEP, CSQA), 3. (TAB COT, GSM8K), and 4. (TAB COT, CSQA). We bold the best result per dataset. Acc represents Accuracy, TNR represents True Negate Rate, TPR represents True Positive Rate (equivalent to recall), Prec represents Precision. SOLAR represents SOLAR 10.7B Instruct; Mistral represents Mistral 7B Instruct v0.2; LLama3 represents Llama 3.0 8B Instruct; Phi3 represents Phi3 Mini 128K Instruct.}
\vspace{-4mm}
\label{table:correctness-classifier-1}
\end{table*}

\subsection{Correctness Classification Task Design}
\label{sec:correctness-pred}
Now that we have presented our setup for the LLM verifier, the question arises on how do we evaluate the capacity of LLMs to self-verify. 
Since we do not have ground-truth labels for correctness of each reasoning step of an entire reasoning chain and because of the difficulty of creating such labels, we design the evaluation of LLMs self-verification as follows: we consider the specific task of correctness classification at the entire reasoning chain level despite our model-based self-verification being at reasoning step-level. 
In this task, the input is a question and the entire reasoning chain for the question. The output is a binary class: $1$ if the reasoning chain has all correct reasoning steps (i.e., all intermediate reasoning steps are predicted as correct by the self-verifier) and the final answer matches the ground truth, and $0$ otherwise. 

Specifically, 
given a question and a reasoning chain as input we decompose it into multiple step-level inputs. A step-level input for step $i$ will have the question and all reasoning steps in the chain from the beginning to step $i$. The LLM is used independently on each step-level input under the self-verification prompt templates (R-prompt, COTR-prompt). For each step-level input we get the LLM's classification on whether the reasoning step is correct or not based on the answer extraction strategy described before ($\S$\ref{sec:eval-answer-extraction}). The final prediction is $1$ (indicating the correctness of the whole chain of reasoning) if the LLM answers affirmatively (option ``(A)") to the correctness of each step-level input for the given reasoning chain and question. Otherwise the final prediction is $0$ (indicating the incorrectness of at least one step in the chain of reasoning). 
Note that for this task, using the verification scores themselves are not necessary. Only the answer extractions are relevant. 
We evaluate the capability of LLMs to self-verify using verification scores in $\S$\ref{sec:RQ3}.


\subsection{Correctness Classification Dataset Design}
Above we described the task of correctness classification for evaluating verification capabilities of LLMs. We also explained how we can use LLMs to predict chain-level correctness or incorrectness based on step-level predictions. However, we still need datasets for the task to actually perform the evaluations. To design the dataset we need three components: (1) the questions (2) the reasoning chain responses to the questions, and (3) the ground truth annotations of the correctness of the reasoning chains for their questions. Below, we explain how we obtain each of these components:

\noindent \textbf{1. Questions:} We get the questions from existing question answering (QA) datasets like GSM8K \cite{cobbe2021training} or CommonsenseQA (CSQA)  \cite{talmor-etal-2019-commonsenseqa}.

\noindent \textbf{2. Reasoning Chains:} Given the questions, we use LLMs with zero-shot COT-style prompts to generate reasoning chains.

\noindent \textbf{3. Ground Truth Annotations:} Given that we have access to QA datasets, for every question, we also have the ground truth answers from the corresponding QA dataset. Given a question, we compare the answer from its generated reasoning chain and the available ground truth answer. While comparing, if the answers match, we consider the reasoning chain to be correct for the question and annotate it as so. Otherwise, we annotate the chain as incorrect.

Based on the differences in source of the questions (different QA datasets) and the differences in the zero-shot prompts for reasoning-chain generation, we create multiple separate datasets. We mainly focus on COT STEP and TAB COT prompts given their ease for thought decomposition. For the source of questions, we consider GSM8K - a mathematical reasoning dataset \cite{cobbe2021training} and CSQA - a commonsense reasoning dataset \cite{talmor-etal-2019-commonsenseqa}. Thus, we get four different datasets based on the combination of prompt style and question source - 1. (COT STEP, GSM8K), 2. (COT STEP, CSQA), 3. (TAB COT, GSM8K), and 4. (TAB COT, CSQA). For each dataset, we run Phi 3 (Phi 3 Mini 128K Instruct) and SOLAR (SOLAR 10.7B Instruct) independently on the first $4,000$ questions from the corresponding QA datasets to generate reasoning chains. For each dataset, from the totality of generated reasoning chains for that dataset, we annotate them automatically based on the described strategy and then subsample it such that there are $1,000$ positive samples and $1,000$ negative samples. Thus we get balanced datasets for evaluating different LLM models on correctness classification for reasoning chains generated by different prompt styles.

\subsection{Self-Verification Results}
We show our results on the four created datasets in Table \ref{table:correctness-classifier-1}. We can observe from the table that when the questions are sourced from the mathematical reasoning dataset (GSM8K), COTR-prompt consistently outperform R-prompt for verification for all the LLM models and reasoning chain prompts that are considered here. The R-prompt leads to a high bias in predicting any given chain as correct. Thus, we find that the R-prompt gets high true positive rate, but extremely low true negative rate. Interestingly, none of the verifier setup performs particularly well when the questions are sourced in CSQA - a commonsense reasoning dataset. This may be because chain of thought-style reasoning is distributionally non-standard for basic commonsense reasoning (which in natural settings, typically happens implicitly).  We also share results with more correction classification datasets created from other zero-shot COT prompts and with other LLMs as verifiers in Appendix \ref{sec:additional}.

\section{RQ3: Utilizing LLM-based Verifier Scores for Reasoning}
\label{sec:RQ3}
We now investigate different ways to utilize the verifier scores to boost performance of LLMs in different reasoning tasks. That is, we analyze how well can LLM-based verifier scores be used in selecting the right answer to a question among many (\S\ref{sec:augmented-sc}) or in guiding reasoning in real-time (\S\ref{vgs}).
\subsection{Verifier Score-based Self-Consistency}
\label{sec:augmented-sc}
Several works \cite{cobbe2021training,li2023making,weng-etal-2023-large, liang2024improving} have tried to utilize verifier scores to augment self-consistency. Similar to the spirit of such prior works, here, we consider three policies to augment self-consistency using verifier scores or verifier predictions. Each of these policies are different strategies to select a single answer from multiple sampled answer. Majority voting is the standard baseline strategy used in typical self-consistency for answer selection. It does not consider any verifier score. We describe below our three alternative policies (that do utilize the verifier score).  For policies (1) and (2), we need the verifier score for an entire reasoning chain. To get that, we simply add up the verifier scores of each step, i.e., the $s_f$ scores from Eq (\ref{eq2}) of each reasoning step (see  $\S$\ref{sec:score-gen}).

\begin{table}[t]
\centering
\small
\def\arraystretch{1.2} 
\begin{tabular}{  l | l l} 
\toprule
\textbf{Prompt Style} & \textbf{GSM8K} & \textbf{CSQA}\\
\hline
\multicolumn{3}{l}{\textbf{SOLAR 10.7B Instruct}}\\
\hline


COT STEP + SC & 84.91 & 79.85\\
COT STEP + SC (Max) & 79.91 & 76.82\\
COT STEP + SC (Weighted) & 84.76 & \textbf{79.94}\\
COT STEP + SC (Filter) & \textbf{84.99} & \textbf{79.94}\\
\hline
\multicolumn{3}{l}{\textbf{Phi3 Mini 128K Instruct}}\\
\hline

COT STEP + SC & \textbf{90.75} & \textbf{77.07}\\
COT STEP + SC (Max) & 87.34 & 73.96\\
COT STEP + SC (Weighted) & 90.68 & 76.66\\
COT STEP + SC (Filter) & 90.37 & 76.90\\
\bottomrule
\end{tabular}
\caption{Comparison of accuracies on GSM8K and CSQA using different policies of utilizing verifier scores in Self-Consistency. SC represents self-consistency with majority voting. SC (Max) represents the use of Max policy on self-consistency. SC (Weighted) represents the use of Weighted policy on self-consistency. SC (Filter) represents the use of Filter policy on self-consistency. In all cases, 20 samples were generated.}
\label{table:filter-or-not}
\end{table}

\noindent \textbf{1. Max:} In this policy, given multiple sampled answers, we choose the one from the highest scoring reasoning chain according to the verifier. 

\noindent \textbf{2. Weighted:} In this policy, given multiple sampled answers, we conduct a weighted majority voting, where the veto power of each sample is the score of its reasoning chain assigned by the verifier. 

\noindent \textbf{3. Filter:} In this policy, given multiple sampled answers, we predict the correctness of each answer using the verifier (based on the correctness prediction strategy described in $\S$\ref{sec:correctness-pred}). We then filter out all the samples that are classified as incorrect and conduct a majority voting on the remaining samples. Note that the answers of samples that are classified as correct can still differ even if the entire reasoning chain is predicted as correct.

\noindent \textbf{Results:} In Table \ref{table:filter-or-not}, we compare all three policies against standard majority voting in two of our main datasets (GSM8K, CSQA) for COT STEP (because it allows better step decomposition for verification). Interestingly, we find that none of the three zero-shot verifier-based policies consistently outperforms the standard majority voting baseline in any significant manner. In Appendix \ref{sec:additional}, we also share similar results with COT TAB prompt. 

\begin{table*}[t]
\small
\centering
\def\arraystretch{1.2} 
\begin{tabular}{  l | l l l l l | l l l l l} 
\toprule
\textbf{Prompt Style} & \textbf{GSM8K} & \textbf{GSMH} & \textbf{AQuA} & \textbf{SQA} & \textbf{CSQA} & \textbf{GSM8K} & \textbf{GSMH} & \textbf{AQuA} & \textbf{SQA} & \textbf{CSQA}\\
\hline
\multicolumn{6}{c|}{\textbf{SOLAR 10.7B Instruct}} & \multicolumn{5}{c}{\textbf{SOLAR 10.7B Instruct + SC@20}}\\
\hline
COT STEP & 73.54 & 30.63 & 11.02 & 58.95 & 67.32 & \textbf{84.91} & \textbf{37.45} & 38.58 & \textbf{73.76} & \textbf{79.85}\\
COT STEP + SG & \textbf{76.50} & \textbf{32.07} & \textbf{15.75} & \textbf{55.85} & \textbf{68.71} & 83.78 & 35.71 & \textbf{42.13} & 73.41 & 79.28\\
\hline
\multicolumn{6}{c|}{\textbf{Phi3 Mini 128K Instruct}} & \multicolumn{5}{c}{\textbf{Phi3 Mini 128K Instruct + SC@20}}\\
\hline
COT STEP & 83.25 & 38.59 & 15.75 & 51 & 71.75 & \textbf{90.75} & 45.49 & 65.75 & 63.49 & 77.07\\
COT STEP + SG & \textbf{85.37} & \textbf{39.65} & \textbf{23.23} & \textbf{52.97} & \textbf{73.2} & 90.30 & \textbf{45.64} & \textbf{67.32} & \textbf{64.54} & \textbf{77.56}\\
\bottomrule
\end{tabular}
\caption{Comparison of accuracy across different datasets. SQA represents StrategyQA. SG represents Step-wise Greedy Search. SC@20 represents Self-consistency with majority voting of 20 sample generations. We bold the best result per block per dataset.}
\label{table:stepwise-greedy}
\end{table*}

\begin{table}[t]
\centering
\small
\def\arraystretch{1.2} 
\begin{tabular}{  l | l l} 
\toprule
\textbf{Prompt Style} & \textbf{GSM8K} & \textbf{CSQA}\\
\hline
\multicolumn{3}{l}{\textbf{SOLAR 11B Instruct}}\\
\hline
COT STEP + SG + SC & \textbf{83.78} & 79.28\\
COT STEP + SB + SC & 78.92 & \textbf{80.75}\\
\hline
\multicolumn{3}{l}{\textbf{Phi3 Mini 128K Instruct}}\\
\hline
COT STEP + SG + SC & \textbf{90.30} & \textbf{77.56}\\
COT STEP + SB + SC & 87.42 & 76.17\\
\bottomrule
\end{tabular}
\caption{Comparison of accuracy from Step-wise Greedy (SG) search and Step-wise Beam (SB) search - both with Self-Consistency (SC) on GSM8K and CSQA (CommonsenseQA). SC  represents Self-consistency with majority voting of 20 sample generations. We bold the best result per block per dataset.}
\label{table:greedy-vs-beam}
\end{table}

\subsection{Verification-Guided Search}
\label{vgs}
In this section, we consider approaches to use the verifier to guide the reasoning steps as they are getting generated rather than just filtering/weighing them post hoc. We provide hyperparameter details about the search methods in Appendix \ref{sec:hyperparameters}. 

\noindent \textbf{Step-wise Greedy Search:} Our main approach is a stochastic Step-wise Greedy Search (SG) similar to \citet{khalifa-etal-2023-grace}. In this strategy, at every turn - first, we use LLMs to generate $k$ reasoning steps (using temperature-based sampling); second, we use the verifier with the COTR-prompt to score each step; and, third, we greedily select the reasoning step with the maximum verification score to condition the generation of the next step. We stop when a reasoning step with the end of sequence marker is generated.  

\noindent \textbf{Step-wise Beam Search:} We also consider an extension of SG where instead of the stochastic step-wise greedy search, we use stochastic Step-wise Beam (SB) search guided by the verifier similar to \citet{xie2023self-evaluation}. The SB approach is also comparable to the breadth-first-search used in Tree of Thoughts \cite{yao2023tree}. We use majority voting among the final beams - we call this overall strategy SB + SC (SC stands for Self-Consistency). We share more technical details in Appendix \ref{sec:hyperparameters}. 

\noindent \textbf{Results:} In Table \ref{table:stepwise-greedy}, we compare basic COT STEP and COT STEP augmented with SG search. Here, we find that our verifier guided SG consistency outperforms the baseline when not using self-consistency (SC). However, when SC is involved for both methods, the difference between the baseline and the more sophisticated SG approach becomes much less substantial. Therefore, for now, SC is still a better trade off for increasing performance than our zero-shot verifier which requires more compute. In Appendix \ref{sec:additional}, we also share comparisons with TAB COT. 

In Table \ref{table:greedy-vs-beam} we show the comparison between SB + SC and SG + SC. We find that SB + SC generally under-performs the simpler SG + SC. This may be because vanilla beam search tends to struggle with diverse generations without additional augmentations \cite{li2016simple}. Our results also suggests the potential of simpler step-level greedy search (SG) as a baseline competitor to consider against other verifier-guided search models using world models or tree of thoughts \cite{hao2023reasoning}.

\section{Related Works}

Multiple works \cite{kadavath2022language,ling2023deductive,stechly2024self,hong-etal-2024-closer} have studies of LLM-based verification and their limitations. However, most of them focus on few-shot verification. Several works studied the potential for LLMs to self-reflect and refine their answers \cite{shin2023reflexion,madaan2023self-refine,paul-etal-2024-refiner,zhou2023solving,zheng2024progressivehint} which involves a form of verification of the initial answer. Contrary to those works, we currently focus on single-round prompting that involves no refinement of the initial answer.
\citet{cobbe2021training,li2023making,weng-etal-2023-large, liang2024improving} trained a separate verifier to score answers or reasoning steps - ultimately to be used in Max or Weighted aggregation policies. \citet{zhou2023don} used an external theorem prover for verification while using LLMs for auto-formalization. \citet{khalifa-etal-2023-grace} used a fine-tuned separate verifier score in a step-wise greedy search. Multiple works \cite{gandhi2023strategic,yao2023tree,xie2023self-evaluation,qiu2024treebon,wan2024alphazerolike,wang2024q,sun2024beats,wang2024towards,gao2024interpretable,besta2024topologies,hao2024llm} explored the utility of using verifier scores as rewards or costs in a tree-search or beam-search approach. However, they either use fine-tuned or few shot-based verifiers. One exception is \citet{miao2024selfcheck} who perform zero-shot self-verification to adjust self-consistency similar to us. However, they use a complicated multi-stage prompting strategy for verification which can become exorbitantly costly when used with tree search approaches \cite{gao2024interpretable}. Moreover, we explore a fully zero-shot regime with zero-shot prompt-based thought decomposition as well. 

\section{Conclusion}
\label{sec:conclusion}
We have the following takeaways: (1) prompts like PS+ or TAB COT are not necessarily systematically better than COT. (2) COT STEP offers an elegant zero-shot strategy to decompose reasoning steps virtually without any accuracy loss compared to COT. (3) Zero-shot COT prompt (COTR-prompt) is also useful for verification particularly in mathematical domain. (4) Zero-shot verifier scores do not particularly help in augmenting self-consistency. (5) Using zero-shot verifier scores to guide reasoning step search in a step-wise stochastic greedy manner can be helpful but its benefit compared to plain COT disappears when using self-consistency. Beam search did not help either. (6) The verifier works better with COT STEP-based decomposition than TAB COT.

\section{Limitations}
We utilize an automatic way of constructing correctness classification dataset. However, this method is not perfect - for it assumes that just because the predicted answer is correct the reasoning chain as a whole is correct. In some cases that may not be true \cite{lanham2023measuring}. However, this is a trade-off to make to get the benefit of automaticity. There exists some data with step-level ground truth verification scores \cite{cobbe2021training,lightman2024lets} however only the automated construction offers a way to conveniently evaluate the verification of different prompts for reasoning step generation. 
  
\bibliography{custom}
\clearpage
\newpage
\appendix
\section{Search Hyperparameters}
\label{sec:hyperparameters}
For step-wise greedy search (SG), at each turn, we generate $k=5$ reasoning steps. When combining SG with SC, at the first turn we generate $20$ reasoning steps. We keep all the $20$ reasoning steps and independently continue $20$ reasoning chains. Each of the $20$ chains use SG as before from the second turn (generating $k=5$ candidates and greedily selecting the maximum scoring candidate). In the end, we get $20$ reasoning chains with $20$ corresponding answers. We then apply majority voting. 

For step-wise beam search (SB), we use a beam width of $16$. That is, at every turn, we generate $16$ reasoning steps per $b$ (let us say) beam (initially $1$). If at least $16$ beams have reached terminal status (i.e. the end of sequence marker is reached by each of the chains in the beams) among the $16b$ beams, we stop the search and return the majority voted result (thus, combining with self-consistency) among all the terminal beams. If at least $16$ beams have not reached a terminal status, we stochastically sample $16$ beams from the $16b$ beams based on their aggregated verifier scores as a pruning step. The sampled $16$ beams become the $b$ beams for the next step.  

Let us say, the verifier score at any step $i$ for some beam $m$ is $s_f^{(i,m)}$ - calculated as described in $\S$\ref{sec:score-gen}. In that case the aggregated verifier score for a beam $m$ generated up to step $j$ would be:
\begin{equation}
sb_{m} = \sum_{i=0}^j \frac{1}{d_m} \log s_f^{(i,m)}
\end{equation}
$sb_m$ represents the mean score for the beam $m$ generated upto the last step so far (in this case, step $j$). $d_m$ represents the depth (number of reasoning steps) of beam $m$.   We maintain the aggregation in logspace. Given multiple $b$ beams, the aggregated score ($sb_{m}$) for any beam $m$ can be normalized into a probability distribution via:
\begin{equation}
    nsb_m = \frac{\exp (sb_m/\tau_m)}{\sum_{i=1}^b exp(sb_i/\tau_i)}
\end{equation}
Thus, we get a probability distribution over all the available beams at a turn, and we can use this for sampling in the pruning step described before. Similar to \citet{xie2023self-evaluation}, we decay the temperature per turn in the beam search by some $\alpha$: $\tau \leftarrow \tau \cdot \alpha$. Following \citet{xie2023self-evaluation}, we set both the initial temperature ($\tau$) and $\alpha$ as $0.5$.

For both SG and SB we set the maximum number of reasoning steps to $18$. We choose the beam width as $16$ to keep close to the self-consistency sampling size ($20$). We keep it less that $20$, for better computational effeciency and also because, due to the lack of pruning in the final turn, the actual number of final candidates can end up as higher than $20$ either way.

\begin{table*}[t]
\centering
\small
\def\arraystretch{1.2} 
\begin{tabular}{  l | l l | l l} 
\toprule
\textbf{Prompt Style} & \textbf{GSM8K} & \textbf{CSQA} & \textbf{GSM8K} & \textbf{CSQA}\\
\hline
\multicolumn{3}{c|}{\textbf{Mistral 7B Instruct v0.2}} & \multicolumn{2}{c}{\textbf{Mistral 7B Instruct v0.2 + SC@20}}\\
\hline
Base & 19.86 & 57.74 & 28.73 & 64.37\\
COT & 49.05 & 61.26 & 68.31 & 71.50\\
PS+ & 45.87 & 54.79 & \textbf{69.83} & \textbf{73.55}\\
TAB COT & 40.11 & 59.05 & 60.8 & 72.07\\
COT STEP & 48.83 & 62.0 & 68.08 & 73.06\\
COT STEP + SG & \textbf{53.83} & \textbf{65.03} & 64.52 & 72.65\\
\hline
\multicolumn{3}{c|}{\textbf{LLama 3.0 8B Instruct}} & \multicolumn{2}{c}{\textbf{LLama 3.0 8B Instruct + SC@20}}\\
\hline
Base & 35.79 & 62.90 & 57.62 & 73.63\\
COT & 73.54 & 70.93 & 84.69 & 75.51\\
PS+ & 60.2 & 39.31 & 86.51 & 74.53\\
TAB COT & 66.34 & 59.21 & 84.23 & 75.02\\
COT STEP & 76.57 & 68.71 & \textbf{87.95} & \textbf{75.68}\\
COT STEP + SG & \textbf{78.54} & \textbf{71.01} & 87.79 & 74.69\\
\bottomrule
\end{tabular}
\caption{Comparison of accuracy from different prompting strategies across GSM8K and CSQA (CommonsenseQA). SC@20 represents Self-consistency with majority voting of 20 sample generations. SG represents step-wise greedy search. We bold the best result per block per dataset.}
\label{table:other-model-results}
\end{table*}

\section{Additional Results}
\label{sec:additional}
In this section, we share several additional experimental results.

\subsection{Zero Shot Prompt Comparisons with Other LLMs}
In Table \ref{table:other-model-results}, we compare the zero shot prompt performance (including step-wise greedy search with COT STEP) of other LLM models like Mistral (Mistral 7B Instruct v0.2) \cite{jiang2023mistral} and Llama3 (LLama 3.0 8B Instruct) \cite{dubey2024llama}. We observe the same patterns that we did with SOLAR 10.7B and Phi3 in the main paper.

\subsection{Evaluating Self-Verification}
In Table \ref{table:correctness-classifier-2}, we present an extended version of Table \ref{table:correctness-classifier-1}. We present the results on more correctness classification datasets generated from reasoning chains of other COT prompts (PS+, COT) and results by more LLM verifiers (Llama3, Mistral). In case of prompts like COT/PS+, we use new-line to decompose reasoning steps. For the mathematical datasets (based on GSM8K), we find that COT STEP reasoning chains lead to the best for most (if not all) LLM models as verifier (with COTR prompt) on balance (Accuracy, F1 measures). This further justifies the choice of COT STEP. R-prompt tends to remain highly biased towards either rejecting any reasoning chain as incorrect (increasing true negative rate at a high cost of true positive rate) or accepting any reasoning chain as correct (increasing true positive rate at a high cost of true negative rate). COTR-prompt, in contrast, tends to have a better balance. For commonsense reasoning-based datasets (based on CommonsenseQA as the question source), our observations remain similar to before as discussed in the main paper. 

\subsection{TAB COT vs COT STEP}
Both TAB COT and COT STEP allows a principled way to automatically parse reasoning steps in a zero-shot regime. As such, we compare these two prompting strategy a bit more closely. First, we check the potential of using verifier scores for adjusting self-consistency aggregations. Table \ref{table:filter-or-not-cot-tab} is an extension of Table \ref{table:filter-or-not} that includes result of TAB COT with the three proposed ($\S$\ref{sec:augmented-sc} verifier score-based aggregation strategies (Max, Weighted, and Filter) to augment self-consistency. The observations are similar to before. Even for TAB COT the majority voting baseline remains a strong contended and neither of the three policies outperform it substantially. However, when comparing COT STEP against TAB COT, COT STEP appears to generally perform better. Second, we also check the potential of using TAB COT decomposition with verifier scores to search the reasoning space via step-wise greedy search (SG) or step-wise beam search. In Table \ref{table:tab-sg-sc-results}, we extend Table \ref{table:stepwise-greedy} with results from TAB COT showing its performance with step-wise greedy search (with and without self-consistency). In Table \ref{table:tab-greedy-vs-beam}, we extend Table \ref{table:greedy-vs-beam} with results from TAB COT showing its performance with step-wise beam search. In both cases, we find COT STEP to perform much better than TAB COT in general.

\subsection{Verification Score Ablation}
In this section we conduct a short ablation study to check whether component C1 (described in $\S$\ref{sec:score-gen} for verifier score computation is important or not. We compare the performance of Phi3 and SOLAR while using them as verifiers with and without C1 for scoring and guiding them through the step-wise greedy search (SG). In Table \ref{table:confidence-matter}, we report the results. The results show that removing C1 generally tend to harm the performance of SG - thus, underscoring the importance of the component. 

\section{Generation Examples}
\label{sec:gen-examples}
We present examples of structured reasoning chain generations based on COT STEP prompt in Figure \ref{fig:example-gen} and some example generations based on COT prompt in Figure \ref{fig:example-gen-2}. As can be seen while the fundamental content in the two cases are not two different, COT prompt do not have any consistent structure or pattern to exploit for zero-shot decomposition, whereas COT STEP prompt cleanly separates reasoning into numbered steps.

\begin{table*}[t]
\centering
\small
\def\arraystretch{1.2} 
\begin{tabular}{  l | l l l l l | l l l l l} 
\toprule
 & \multicolumn{5}{c|}{\textbf{Question Source: GSM8K}} & \multicolumn{5}{c}{\textbf{Question Source: CSQA}}\\
\textbf{Verifier Setup} & \textbf{Acc} & \textbf{TNR} & \textbf{TPR} & \textbf{Prec} & \textbf{F$_1$} & \textbf{Acc} & \textbf{TNR} & \textbf{TPR} & \textbf{Prec} & \textbf{F$_1$}\\
\hline
\multicolumn{11}{l}{\textbf{Reasoning Chain Generation Prompt: COT}}\\
\hline
SOLAR + R-prompt &  56.10 & 13.8 & \textbf{98.4} & 53.30 & 69.15 & 52.2 & 10.7 & \textbf{93.7} & 51.20 & \textbf{66.22}\\
SOLAR + COTR-prompt & \textbf{78.55} & 76.9 & 80.2 & 77.64 & \textbf{78.90} & 52.30 & 25.5 & 79.10 & 51.50 & 62.40\\
Mistral + R-prompt & 61.45 & 84.0 & 38.9 & 70.86 & 50.23 & \textbf{54.90} & 61.4 & 48.4 & 55.63 & 51.77\\
Mistral + COTR-prompt & 62.55 & 84.8 & 40.30 & 72.61 & 51.8 & 51.5 & 59.1 & 43.9 & 51.77 & 47.51\\
Llama3 + R-prompt & 50.1 & \textbf{99.9} & 0.3 & 75.0 & 0.60 & 54.05 & \textbf{94.1} & 14.00 & \textbf{70.35} & 23.35\\
LLama3 + COTR-prompt & 54.75 & 99.3 & 10.2 & \textbf{93.60} & 18.40 & 54.45 & 88.2 & 20.7 & 63.70 & 31.30\\
Phi3 + R-prompt & 61.35 & 26.5 & 96.2 & 56.69 & 71.34 & 51.6 & 58.9 & 44.3 & 51.87 & 47.79\\
Phi3 + COTR-prompt & 76.35 & 68.0 & 84.7 & 72.58 & 78.17 & 51.45 & 9.8 & 93.10 & 50.79 & 65.73\\
\hline
\multicolumn{11}{l}{\textbf{Reasoning Chain Generation Prompt: PS+}}\\
\hline
SOLAR + R-prompt & 63.25 & 69.90 & 56.60 & 65.28 & 60.63 & 51.80 & 28.9 & 74.7 & 51.24 & 60.78\\
SOLAR + COTR-prompt & 56.90 & 15.30 & \textbf{98.5} & 53.77 & 69.56 & \textbf{54.35} & 57.60 & 51.1 & \textbf{54.65} & 52.82\\
Mistral + R-prompt & 55.3 & 95.20 & 15.4 & 76.24 & 25.62 & 52.55 & 72.3 & 32.80 & 54.22 & 40.87\\
Mistral + COTR-prompt & 60.15 & 89.0 & 31.3 & 74 & 44 & 50.6 & 68.7 & 32.5 & 50.94 & 39.68\\
LLama3 + R-prompt & 50.0 & \textbf{100.0} & 0.0 & 0 & 0 & 50.15 & \textbf{99.3} & 1.0 & 58.82 & 1.97\\
Llama3 + COTR-prompt & 50.5 & 99.8 & 1.2 & \textbf{85.71} & 2.37 & 51.5 & 94.0 & 9.0 & 60.0 & 15.65\\
Phi3 + R-prompt & 62.25 & 32.4 & 92.10 & 57.67 & 70.93 & 50.8 & 73.1 & 28.5 & 51.44 & 36.68\\
Phi3 + COTR -prompt & \textbf{78.25} & 70.3 & 86.2 & 74.38 & \textbf{79.85} & 51.85 & 10.2 & \textbf{93.5} & 51.01 & \textbf{66.01}\\
\hline
\multicolumn{11}{l}{\textbf{Reasoning Chain Generation Prompt: TAB COT}}\\
\hline
SOLAR + R-prompt & 57.15 & 16.9 & 97.40 & 53.96 & 69.45 & 52.85 & 9.6 & \textbf{96.1} & 51.53 & \textbf{67.09}\\
SOLAR + COTR-prompt & 67.30 & 49.8 & 84.8 & 62.82 & 72.17 & 53.55 & 30.60 & 76.5 & 52.43 & 62.22\\
Mistral + R-prompt & 65.3 & 72.2  & 58.4 & 67.75 & 62.73 & \textbf{54.35} & 44.2 & 64.5 & 53.62 & 58.56\\
Mistral + COTR-prompt & 64.15 & 88.8 & 39.5 & \textbf{77.91} & 52.42 & 51.1 & 60.8 & 41.4 & 51.37 & 45.85\\
Llama3 + R-prompt & 50.0 & \textbf{100.0} & 0.0 & 0 & 0 & 50.3 & \textbf{99.7} & 0.90 & \textbf{75.0} & 1.78\\
Llama3 + COTR-prompt & 54.9 & 99.2 & 10.6 & 92.98 & 19.03 & 52.95 & 91.60 & 14.30 & 63 & 23.31\\
Phi3 + R-prompt & 61.35 & 28.7 & 94.0 & 56.87 & 70.86 & 49.4 & 56.90 & 41.9 & 49.29 & 45.30\\
Phi3 + COTR-prompt & \textbf{80.2} & 73.5 & 86.9 & 76.63 & \textbf{81.44} & 51.95 & 15.0 & 88.9 & 51.12 & 64.91\\
\hline
\multicolumn{11}{l}{\textbf{Reasoning Chain Generation Prompt: COT STEP}}\\
\hline
SOLAR + R-prompt & 55.75 & 12.7 & \textbf{98.8} & 53.09 & 69.07 & 50.2 & 1.7 & \textbf{98.7} & 50.10 & 66.47\\
SOLAR + COTR-prompt & \textbf{83.90} & 74.7 & 93.10 & \textbf{78.63} & \textbf{85.26} & 49.5 & 29.80 & 69.20 & 49.64 & 57.81\\
Mistral + R-prompt & 65.3 & 68.30 & 62.3 & 66.28 & 64.23 & 51.2 & 32.30 & \textbf{70.1} & 50.87 & \textbf{58.96}\\
Mistral + COTR-prompt & 68.4 & 72.3 & 64.5 & 69.96 & 67.12 & 49.25 & 41.6 & 56.90 & 49.35 & 52.86\\
Llama3 + R-prompt & 50.35 & \textbf{99.9} & 0.8 & 88.89 & 1.60 & 52.55 & 93.10 & 12.0 & 63.50 & 20.19\\
Llama3 + COTR-prompt & 60.70 & 97.0 & 24.4 & 89.05 & 38.31 & 53.05 & 82.1 & \textbf{24.0} & 57.28 & \textbf{33.83}\\
Phi3 + R-prompt & 59.8 & 22.7 & 96.90 & 55.63 & 70.68 & 45.4 & 57.50 & 33.30 & 43.93 & 37.88\\
Phi3 + COTR-prompt & 75.2 & 60.20 & 90.2 & 69.40 & 78.44 & 50.4 & 10.2 & 90.60 & 50.22 & 64.62\\
\bottomrule
\end{tabular}
\caption{Performance of different LLM models on eight correctness classification datasets created based on four COT-based zero shot prompts and two QA datasets - 1. (COT STEP, GSM8K), 2. (COT STEP, CSQA), 3. (TAB COT, GSM8K), 4. (TAB COT, CSQA) 5. (COT, GSM8K), 6. (COT, CSQA), 7. (PS+, GSM8K), and 8. (PS+, CSQA). We bold the best result per dataset. Acc represents Accuracy, TNR represents True Negate Rate, TPR represents True Positive Rate (equivalent to recall), Prec represents Precision. SOLAR represents SOLAR 10.7B Instruct; Mistral represents Mistral 7B Instruct v0.2; LLama3 represents Llama 3.0 8B Instruct; Phi3 represents Phi3 Mini 128K Instruct.}
\label{table:correctness-classifier-2}
\end{table*}

\begin{table}[t]
\centering
\small
\def\arraystretch{1.2} 
\begin{tabular}{  l | l l} 
\toprule
\textbf{Prompt Style} & \textbf{GSM8K} & \textbf{CSQA}\\
\hline
\multicolumn{3}{l}{\textbf{SOLAR 10.7B Instruct}}\\
\hline
TAB COT + SC & 80.52 & \textbf{80.51}\\
TAB COT + SC (Max) & 72.56 & 75.52\\
TAB COT + SC (Weighted) & 80.82 & 80.84\\
TAB COT + SC (Filter) & 80.21 & 80.34\\
\hline
COT STEP + SC & 84.91 & 79.85\\
COT STEP + SC (Max) & 79.91 & 76.82\\
COT STEP + SC (Weighted) & 84.76 & 79.94\\
COT STEP + SC (Filter) & \textbf{84.99} & 79.94\\
\hline
\multicolumn{3}{l}{\textbf{Phi3 Mini 128K Instruct}}\\
\hline
TAB COT + SC & 86.81 & 74.86\\
TAB COT + SC (Max) & 83.85 & 71.83\\
TAB COT + SC (Weighted) & 87.26 & 74.94\\
TAB COT + SC (Filter) & 86.51 & 74.86\\
\hline
COT STEP + SC & \textbf{90.75} & \textbf{77.07}\\
COT STEP + SC (Max) & 87.34 & 73.96\\
COT STEP + SC (Weighted) & 90.68 & 76.66\\
COT STEP + SC (Filter) & 90.37 & 76.90\\
\bottomrule
\end{tabular}
\caption{Comparison of accuracies on GSM8K and CSQA using different policies of utilizing verifier scores in Self-Consistency. SC represents self-consistency with majority voting. SC (Max) represents the use of Max policy on self-consistency. SC (Weighted) represents the use of Weighted policy on self-consistency. SC (Filter) represents the use of Filter policy on self-consistency. In all cases, 20 samples were generated.}
\label{table:filter-or-not-cot-tab}
\end{table}

\begin{table}[t]
\centering
\small
\def\arraystretch{1.2} 
\begin{tabular}{  l | l l} 
\toprule
\textbf{Prompt Style} & \textbf{GSM8K} & \textbf{CSQA}\\
\hline
\multicolumn{3}{l}{\textbf{SOLAR 10.7B Instruct}}\\
\hline
TAB COT + SG & 70.13 & 67.90\\
TAB COT + SG + SC & 75.51 & 77.72\\
COT STEP + SG & 76.50 & 68.71\\
COT STEP + SG + SC & \textbf{83.78} & \textbf{79.28}\\
\hline
\multicolumn{3}{l}{\textbf{Phi3 Mini 128K Instruct}}\\
\hline
TAB COT + SG & 79.30 & 70.19\\
TAB COT + SG + SC & 79.30 & 72.81\\
COT STEP + SG & 85.37 & 73.22\\
COT STEP + SG + SC & \textbf{90.30} & \textbf{77.56}\\
\bottomrule
\end{tabular}
\caption{Comparison of accuracy from Step-wise Greedy (SG) with and without Self-Consistency (SC) on GSM8K and CSQA (CommonsenseQA). SC  represents Self-consistency with majority voting of 20 sample generations. We bold the best result per block per dataset.}
\label{table:tab-sg-sc-results}
\end{table}

\begin{table}[t]
\centering
\small
\def\arraystretch{1.2} 
\begin{tabular}{  l | l l} 
\toprule
\textbf{Prompt Style} & \textbf{GSM8K} & \textbf{CSQA}\\
\hline
\multicolumn{3}{l}{\textbf{SOLAR 11B Instruct}}\\
\hline
TAB COT + SG + SC & 80.52 & 80.51\\
TAB COT + SB + SC & 70.08 & 77.40\\
COT STEP + SG + SC & \textbf{83.78} & \textbf{79.28}\\
COT STEP + SB + SC & 78.92 & 80.75\\
\hline
\multicolumn{3}{l}{\textbf{Phi3 Mini 128K Instruct}}\\
\hline
TAB COT + SG + SC & 79.30 & 72.81\\
TAB COT + SB + SC & 81.20 & 71.75\\
COT STEP + SG + SC & \textbf{90.30} & \textbf{77.56}\\
COT STEP + SB + SC & 87.42 & 76.17\\
\bottomrule
\end{tabular}
\caption{Comparison of accuracy from Step-wise Greedy (SG) search and Step-wise Beam (SB) search - both with Self-Consistency (SC) on GSM8K and CSQA (CommonsenseQA). SC  represents Self-consistency with majority voting of 20 sample generations. We bold the best result per block per dataset.}
\label{table:tab-greedy-vs-beam}
\end{table}

\begin{table}[t]
\centering
\small
\def\arraystretch{1.2} 
\begin{tabular}{  l | l l} 
\toprule
\textbf{Prompt Style} & \textbf{GSM8K} & \textbf{CSQA}\\
\hline
\multicolumn{3}{l}{\textbf{SOLAR 11B Instruct}}\\
\hline
COT STEP + SG &  76.5 & 68.71\\
COT STEP + SG - C1 & 74.91 & 66.18\\
COT STEP + SG + SC & \textbf{83.78} & 79.28\\
COT STEP + SG + SC - C1 & 82.11 & \textbf{80.10}\\
\hline
\multicolumn{3}{l}{\textbf{Phi3 Mini 128K Instruct}}\\
\hline
COT STEP + SG & 85.37 & 73.22\\
COT STEP + SG - C1 & 84.38 & 72.56\\
COT STEP + SG + SC & \textbf{90.30} & \textbf{77.56}\\
COT STEP + SG + SC - C1 & 90.07 & 75.92\\
\bottomrule
\end{tabular}
\caption{Accuracy comparison under step-wise greedy search with and without C1 component in verifier score computation. $-$C1 represents that the component C1 for verifier score calculation is not used. SG represents step-wise greedy search. SC represents self-consistency with $20$ samples. We bold the best results per block.}
\label{table:confidence-matter}
\end{table}

\begin{figure*}[ht]
\small
\fbox{\parbox{0.985\textwidth}{
\textbf{User:}
Q: Every day, Wendi feeds each of her chickens three cups of mixed chicken feed, containing seeds, mealworms and vegetables to help keep them healthy.  She gives the chickens their feed in three separate meals. In the morning, she gives her flock of chickens 15 cups of feed.  In the afternoon, she gives her chickens another 25 cups of feed.  How many cups of feed does she need to give her chickens in the final meal of the day if the size of Wendi's flock is 20 chickens?<|end|>

\textbf{Assistant:}

A: Let's think step by step.

\textbf{Step 1:}  

First, we need to find out the total amount of feed given to the chickens in the morning and afternoon. 

Morning feed: 15 cups
Afternoon feed: 25 cups

Total feed given in the morning and afternoon: 15 cups + 25 cups = 40 cups

\textbf{Step 2:} 

Now, we know that each chicken gets 3 cups of feed per day. Since Wendi has 20 chickens, we can calculate the total amount of feed needed for all chickens in one day.

Total feed needed for all chickens: 20 chickens * 3 cups/chicken = 60 cups

\textbf{Step 3:}

Now, we can find out how much feed is left for the final meal after accounting for the morning and afternoon feed.

Feed left for the final meal: Total daily feed needed (60 cups) - Feed given in the morning and afternoon (40 cups) = 20 cups

So, Wendi needs to give her chickens 20 cups of feed in the final meal of the day.
Therefore, the answer is 20 cups.

\par\noindent\rule{\textwidth}{0.4pt}

\textbf{User:}
Q: Toulouse has twice as many sheep as Charleston. Charleston has 4 times as many sheep as Seattle. How many sheep do Toulouse, Charleston, and Seattle have together if Seattle has 20 sheep?

\textbf{Assistant:}

A: Let's think step by step.

\textbf{Step 1:}

Seattle has 20 sheep.

\textbf{Step 2:}

Charleston has 4 times as many sheep as Seattle, so Charleston has 4 * 20 = 80 sheep.

\textbf{Step 3:}

Toulouse has twice as many sheep as Charleston, so Toulouse has 2 * 80 = 160 sheep.

\textbf{Step 4:} 

To find the total number of sheep, we add the number of sheep in each city: 20 (Seattle) + 80 (Charleston) + 160 (Toulouse) = 260 sheep.

So, Toulouse, Charleston, and Seattle have a total of 260 sheep together.
Therefore, the answer is 260.

}}
\caption{Example generations by Phi 3 Mini Instruct based on COT STEP prompt on GSM8K samples. Bold is applied for stylistic reasons.}
\label{fig:example-gen}
\end{figure*}
\begin{figure*}[ht]
\small
\fbox{\parbox{0.985\textwidth}{
\textbf{User:}
Q: Every day, Wendi feeds each of her chickens three cups of mixed chicken feed, containing seeds, mealworms and vegetables to help keep them healthy.  She gives the chickens their feed in three separate meals. In the morning, she gives her flock of chickens 15 cups of feed.  In the afternoon, she gives her chickens another 25 cups of feed.  How many cups of feed does she need to give her chickens in the final meal of the day if the size of Wendi's flock is 20 chickens?

\textbf{Assistant:}

A: Let's think step by step.

First, we know that Wendi feeds each chicken 3 cups of feed per day in total. 

So, for 20 chickens, the total feed required would be:
20 chickens * 3 cups/chicken = 60 cups

Now, let's see how much feed Wendi has already given in the morning and the afternoon:

Morning feed = 15 cups
Afternoon feed = 25 cups

Total feed given so far = 15 cups + 25 cups = 40 cups

Now, let's subtract the feed given so far from the total daily requirement to find out how much feed is needed for the final meal:

Feed needed for the final meal = Total daily requirement - Feed given so far
Feed needed for the final meal = 60 cups - 40 cups = 20 cups

Therefore, Wendi needs to give her chickens 20 cups of feed in the final meal of the day.
Therefore, the answer is 20 cups.

\par\noindent\rule{\textwidth}{0.4pt}

\textbf{User:}
Q: Toulouse has twice as many sheep as Charleston. Charleston has 4 times as many sheep as Seattle. How many sheep do Toulouse, Charleston, and Seattle have together if Seattle has 20 sheep?

\textbf{Assistant:}

A: Let's think step by step.

1. Seattle has 20 sheep.
2. Charleston has 4 times as many sheep as Seattle, so Charleston has 4 * 20 = 80 sheep.
3. Toulouse has twice as many sheep as Charleston, so Toulouse has 2 * 80 = 160 sheep.

Now, to find the total number of sheep all three cities have together, we add the number of sheep from each city:

Seattle: 20 sheep
Charleston: 80 sheep
Toulouse: 160 sheep

Total = 20 + 80 + 160 = 260 sheep.
Therefore, the answer is 260 sheep.

}}
\caption{Example generations by Phi 3 Mini Instruct based on COT prompt on GSM8K samples. Bold is applied for stylistic reasons.}
\label{fig:example-gen-2}
\end{figure*}


\end{document}